#nr14 **Short abstract**

**Info-computational constructivism in modelling of life as cognition**

Gordana Dodig Crnkovic
Computer Science Department, Mälardalen University, Västerås, Sweden;
E-Mail: gordana.dodig-crnkovic@mdh.se

This paper addresses the open question formulated as: "Which levels of abstraction are appropriate in the synthetic modelling of life and cognition?" within the framework of info-computational constructivism, treating natural phenomena as computational processes on informational structures.

At present we lack the common understanding of the processes of life and cognition in living organisms with the details of co-construction of informational structures and computational processes in embodied, embedded cognizing agents, both living and artifactual ones.

Starting with the definition of an agent as an entity capable of acting on its own behalf, as an actor in Hewitt's Actor model of computation, even so simple systems as molecules can be modelled as actors exchanging messages (information). We adopt Kauffman's view of a living agent as something that can reproduce and undergoes at least one thermodynamic work cycle. This definition of living agents leads to the Maturana and Varela's identification of life with cognition.

Within the info-computational constructive approach to living beings as cognizing agents, from the simplest to the most complex living systems, mechanisms of cognition can be studied in order to construct synthetic model classes of artifactual cognizing agents on different levels of organization.

#nr14 **Long abstract**

**INFO-COMPUTATIONAL CONSTRUCTIVISM IN MODELLING OF LIFE AS COGNITION**

Gordana Dodig Crnkovic

Computer Science Department, Mälardalen University, Västerås, Sweden;

E-Mail: gordana.dodig-crnkovic@mdh.se

**Life as info-computational generative process of cognition at different levels of organization**

This paper presents a study within info-computational constructive framework of the life process as <knowledge> generation in living agents from the simplest living organisms to the most complex ones. Here <knowledge> of a primitive life form is very basic indeed – it is <knowledge> how to act in the world. An amoeba <knows> how to search for food and how to avoid dangers.

An agent is defined as an entity capable of acting on its own behalf. It can be seen as an "actor" in the Actor model of computation in which "actors" are the basic elements of concurrent computation exchanging messages, capable of making local decisions and creating new actors. Computation is thus distributed in space where computational units communicate asynchronously and the entire computation is not in any well-defined state. (An actor can have information about other actors that it has received in a message about what it was like when the message was sent.) (Hewitt, 2012)

A living agent is a special kind of actor that can reproduce and that *undergoes at least one thermodynamic work cycle*. (Kauffman, 2000) This definition differs from the common belief that (living) agency requires beliefs and desires, unless we ascribe some primitive form of <belief> and <desire> even to a very simple living agents such as bacteria. The fact is that they act on some kind of <anticipation> and according to some <preferences> which might be automatic in a sense that they directly derive from the organisms morphology. Even the simplest living beings act on their own behalf.

Although a detailed physical account of the agents capacity to perform work cycles and so persist in the world is central for understanding of life/cognition, as (Kauffman, 2000) (Deacon, 2007) have argued in detail, this paper will be primarily interested of the info-computational aspects of life. Given that there is no information without physical implementation (Landauer, 1991), computation as the dynamics of information is the *execution of physical laws*.

Kauffman's concept of agency (also adopted by Deacon) suggests the possibility that *life can be derived from physics*. That is not the same as to claim that *life can be reduced to physics* that is obviously false. However, in *deriving life from physics* one may expect that both our understanding of life as well as physics will change. We witness the emergence of information physics (Goyal, 2012) (Chiribella, G.; D'Ariano, G.M.; Perinotti, 2012) as a possible reformulation of physics that may bring physics and life/cognition closer to each other. This development smoothly connects to info-computational understanding of nature (Dodig-Crnkovic & Giovagnoli, 2013).

Life can be analyzed as cognitive processes unfolding in a layered structure of nested information network hierarchies with corresponding computational dynamics (information processes) – from molecular, to cellular, organismic and social levels.

**The computing nature**

This section will introduce two fundamental theories about the nature of the universe and propose their synthesis. The first one with focus on *processes* is the idea of *computing universe* (naturalist computationalism/ pancomputationalism) in which one sees the dynamics of physical states in nature as information processing (natural computation).

The parallel fundamental theory with focus on *structures* is *Informational structural realism* (Floridi, 2003) that takes information to be the fabric of the universe (for an agent). Following definitions of (Hewitt, 2007) and (Bateson, 1972) *information is defined as the difference in one physical system that makes the difference in another physical system*. Of special interest with respect to <knowledge> generation are *agents - systems able to act on their own behalf* and make sense (use) of information.
This relates to the ideas of participatory universe, (Wheeler, 1990) endophysics (Rössler, 1998) and observer-dependent <knowledge> production.

As a synthesis of informational structural realism and natural computationalism, info-computational structuralism adopts two basic concepts: information (as a structure) and computation (as a dynamics of an informational structure) (Dodig-Crnkovic, 2011) (Chaitin, 2007). In consequence the process of dynamical changes of the universe makes the universe a huge computational network where computation is information processing. (Dodig-Crnkovic & Giovagnoli, 2013) Information and computation are two basic and inseparable elements necessary for naturalizing cognition and <knowledge>. (Dodig-Crnkovic, 2009)

**<Knowledge> generation as morphological computation – from simplest to the most complex organisms**

In the computing nature, <knowledge> generation should be studied as a natural process. That is the main idea of Naturalized epistemology (Harms, 2006), where the subject matter is not our concept of <knowledge>, but the knowledge itself as it appears in the world as specific informational structures of an agent. Maturana was the first to suggest that *knowledge is a biological phenomenon*. He and Varela argued that life should be understood as a process of cognition, which enables an organism to adapt and survive in the changing environment. (Maturana & Varela, 1980)

The origin of <knowledge> in first living agents is not well researched, as the idea still prevails that only humans possess knowledge. However, there are different types of <knowledge> and we have good reasons to ascribe "knowledge how" and even simpler kinds of "knowledge that" to other living beings. Plants can be said to possess memory (in their bodily structures) and ability to learn (adapt, change their morphology) and can be argued to possess rudimentary forms of knowledge. On the topic of plant cognition see Garzón in (Pombo, O., Torres J.M., Symons J., 2012) p. 121. In his *Anticipatory systems* (Rosen, 1985) as well as Popper in (Popper, 1999) p. 61 ascribe the ability to <know> to all living: "Obviously, in the biological and evolutionary sense in which I speak of knowledge, not only animals and men have expectations and therefore (unconscious) knowledge, but also plants; and, indeed, all organisms."

Computation as information processing should not be identified with classical cognitive science, with the related notions of input–output and structural representations – but it is important to recognize that *also connectionist models are computational* as they are also based on information processing (Scheutz, 2002)(Dodig-Crnkovic, 2009). The basis for the capacity to acquire <knowledge> is in the specific morphology of organisms that enables perception, memory and adequate information processing that can lead to production of new <knowledge> out of old one. Harms proved a theorem showing that natural selection will always lead a population to accumulate information, and so to 'learn' about its environment: "any evolving population 'learns' about its environment, in Harms' sense, even if the population is composed of organisms that lack minds entirely, hence lack the ability to have representations of the external world at all. " (Okasha, 2005)

In this context, open vs. closed system modelling (Burgin & Dodig-Crnkovic, 2013) is addressed. In order to develop general theory of the networked physical information processing, we must also generalize the ideas of what computation is and what it might be. For new computing paradigms, see for example (Rozenberg, Bäck, & Kok, 2012)(Burgin, 2005)(MacLennan, 2004) (Wegner, 1998)(Hewitt, 2012)(Abramsky, 2008).

**Summary**

This article argues that *computational modelling of self-organization and autopoiesis (as a special case of self-generative processes in living organism) are fundamentally info-computational processes* of morphological computing that develop at different levels of organization of physical systems.
The following are the main points:

1. Info-computational constructivism as a framework for the analysis:

- *Information*: Reality as information for an agent. Information is observer-dependent.

- *Computation*: natural/morphological computation, self-organization of information. Execution of physical laws on different levels of organization. <Knowledge> production as computation.

2. Agency as understood within the Actor model of asynchronous computation.

3. For an actor to qualify as living agent, it must be capable of sustaining at least one thermodynamic work cycle.

4. Life as a (morphological computational) process of cognition. On each level of organization corresponding level of cognition is present, from single cell to an organism and society (social cognition).

5. Knowledge generation as morphological computation of informational structures.